\documentclass{article}

\usepackage[preprint]{Styles/neurips_2025}

\newif\ifanonsubmission
\anonsubmissionfalse
\ifanonsubmission
  \newcommand{\chigeom}{\textsc{XYZ-Lib}}
\else
  \newcommand{\chigeom}{Chi-Geometry}
\fi

\newif\ifresolutionsubmission
\resolutionsubmissionfalse

\usepackage[utf8]{inputenc} 
\usepackage[T1]{fontenc}    
\usepackage{hyperref}       
\usepackage{url}            
\usepackage{booktabs}       
\usepackage{amsfonts}       
\usepackage{nicefrac}       
\usepackage{microtype}      
\usepackage{xcolor}         
\usepackage{acro}
\usepackage{graphicx}
\usepackage{subcaption}
\usepackage{booktabs}
\usepackage{tabularx}
\usepackage{footnote}
\usepackage{colortbl}
\usepackage{amsmath}
\usepackage{amsfonts}
\usepackage{upgreek}

\title{\chigeom{}: A Library for Benchmarking Chirality Prediction of GNNs}

\author{%
  Rylie Weaver\\
  Computational Sciences and Engineering Division\\
  Oak Ridge National Laboratory\\
  Oak Ridge, TN 37830 \\
  \texttt{weaverre@ornl.gov} \\
  \And
  Massamiliano Lupo Pasini\\
  Computational Sciences and Engineering Division\\
  Oak Ridge National Laboratory\\
  Oak Ridge, TN 37830 \\
  \texttt{lupopasinim@ornl.gov} \\
}

\DeclareAcronym{ai}{
    short = AI ,
    long  = Artificial Intelligence
}

\DeclareAcronym{cip}{
    short = CIP ,
    long  = Cahn-Ingold Prelog
}

\DeclareAcronym{dl}{
    short = DL ,
    long  = Deep Learning
}

\DeclareAcronym{ecd}{
    short = ECD ,
    long  = electronic circular dichroism
}

\DeclareAcronym{gnn}{
  short = GNN ,
  long  = graph neural network
}

\DeclareAcronym{mpnn}{
    short = MPNN ,
    long  = message passing neural network
}

\DeclareAcronym{ml}{
    short = ML ,
    long  = Machine Learning
}

\DeclareAcronym{stp}{
    short = STP ,
    long  = scalar triple product
}

\DeclareAcronym{sota}{
    short = SOTA ,
    long  = state-of-the-art
}

\begin{document}
\maketitle

\ifresolutionsubmission
    \begingroup
    \renewcommand\thefootnote{}     
    
    \footnotetext{%
        \textbf{Notice of Copyright and Acknowledgements.}\,
        This manuscript has been authored by UT-Battelle, LLC, under contract DE-AC05-00OR22725
        with the US Department of Energy (DOE). The US government retains and the publisher, by
        accepting the article for publication, acknowledges that the US government retains a
        nonexclusive, paid-up, irrevocable, worldwide license to publish or reproduce the published form of this manuscript, or allow others to do so, for US government purposes. DOE will provide public access to these results of federally sponsored research in accordance with the DOE Public Access Plan (http://energy.gov/downloads/doe-public-access-plan). This manuscript is produced by research sponsored by the Artificial Intelligence Initiative as part of the Laboratory Directed Research and Development (LDRD) Program of Oak Ridge National Laboratory (ORNL), supported in part by an appointment to the ORNL Research Student Internships Program, sponsored by the US DOE and administered by the Oak Ridge Institute for Science and Education.}
    \addtocounter{footnote}{-1}
    \endgroup
\fi

\definecolor{my100}{HTML}{00A500}
\definecolor{my99}{HTML}{05A500}
\definecolor{my98}{HTML}{0AA500}
\definecolor{my97}{HTML}{0FA500}
\definecolor{my96}{HTML}{14A500}
\definecolor{my95}{HTML}{19A500}
\definecolor{my94}{HTML}{1EA500}
\definecolor{my93}{HTML}{23A500}
\definecolor{my92}{HTML}{28A500}
\definecolor{my91}{HTML}{2DA500}
\definecolor{my90}{HTML}{32A500}
\definecolor{my89}{HTML}{38A500}
\definecolor{my88}{HTML}{3DA500}
\definecolor{my87}{HTML}{42A500}
\definecolor{my86}{HTML}{47A500}
\definecolor{my85}{HTML}{4CA500}
\definecolor{my84}{HTML}{51A500}
\definecolor{my83}{HTML}{56A500}
\definecolor{my82}{HTML}{5BA500}
\definecolor{my81}{HTML}{60A500}
\definecolor{my80}{HTML}{66A500}
\definecolor{my79}{HTML}{6BA500}
\definecolor{my78}{HTML}{70A500}
\definecolor{my77}{HTML}{75A500}
\definecolor{my76}{HTML}{7AA500}
\definecolor{my75}{HTML}{7FA500}
\definecolor{my74}{HTML}{84A500}
\definecolor{my73}{HTML}{89A500}
\definecolor{my72}{HTML}{8EA500}
\definecolor{my71}{HTML}{93A500}
\definecolor{my70}{HTML}{99A500}
\definecolor{my69}{HTML}{9EA500}
\definecolor{my68}{HTML}{A3A500}
\definecolor{my67}{HTML}{A8A500}
\definecolor{my66}{HTML}{ADA500}
\definecolor{my65}{HTML}{B2A500}
\definecolor{my64}{HTML}{B7A500}
\definecolor{my63}{HTML}{BCA500}
\definecolor{my62}{HTML}{C1A500}
\definecolor{my61}{HTML}{C6A500}
\definecolor{my60}{HTML}{CCA500}
\definecolor{my59}{HTML}{D1A500}
\definecolor{my58}{HTML}{D6A500}
\definecolor{my57}{HTML}{DBA500}
\definecolor{my56}{HTML}{E0A500}
\definecolor{my55}{HTML}{E5A500}
\definecolor{my54}{HTML}{EAA500}
\definecolor{my53}{HTML}{EFA500}
\definecolor{my52}{HTML}{F4A500}
\definecolor{my51}{HTML}{F9A500}
\definecolor{my50}{HTML}{FFA500}
\definecolor{my49}{HTML}{FC9F00}
\definecolor{my48}{HTML}{FA9800}
\definecolor{my47}{HTML}{F89100}
\definecolor{my46}{HTML}{F68B00}
\definecolor{my45}{HTML}{F48400}
\definecolor{my44}{HTML}{F27D00}
\definecolor{my43}{HTML}{F07700}
\definecolor{my42}{HTML}{EE7000}
\definecolor{my41}{HTML}{EC6A00}
\definecolor{my40}{HTML}{EA6300}
\definecolor{my39}{HTML}{E85C00}
\definecolor{my38}{HTML}{E65600}
\definecolor{my37}{HTML}{E44F00}
\definecolor{my36}{HTML}{E24800}
\definecolor{my35}{HTML}{E04200}
\definecolor{my34}{HTML}{DE3B00}
\definecolor{my33}{HTML}{DC3500}
\definecolor{my32}{HTML}{DA2E00}
\definecolor{my31}{HTML}{D82700}
\definecolor{my30}{HTML}{D62100}
\definecolor{my29}{HTML}{D41A00}
\definecolor{my28}{HTML}{D21300}
\definecolor{my27}{HTML}{D00D00}
\definecolor{my26}{HTML}{CE0600}
\definecolor{my25}{HTML}{CC0000}
\definecolor{my24}{HTML}{CC0000}
\definecolor{my23}{HTML}{CC0000}
\definecolor{my22}{HTML}{CC0000}
\definecolor{my21}{HTML}{CC0000}
\definecolor{my20}{HTML}{CC0000}
\definecolor{my19}{HTML}{CC0000}
\definecolor{my18}{HTML}{CC0000}
\definecolor{my17}{HTML}{CC0000}
\definecolor{my16}{HTML}{CC0000}
\definecolor{my15}{HTML}{CC0000}
\definecolor{my14}{HTML}{CC0000}
\definecolor{my13}{HTML}{CC0000}
\definecolor{my12}{HTML}{CC0000}
\definecolor{my11}{HTML}{CC0000}
\definecolor{my10}{HTML}{CC0000}
\definecolor{my9}{HTML}{CC0000}
\definecolor{my8}{HTML}{CC0000}
\definecolor{my7}{HTML}{CC0000}
\definecolor{my6}{HTML}{CC0000}
\definecolor{my5}{HTML}{CC0000}
\definecolor{my4}{HTML}{CC0000}
\definecolor{my3}{HTML}{CC0000}
\definecolor{my2}{HTML}{CC0000}
\definecolor{my1}{HTML}{CC0000}
\definecolor{my0}{HTML}{CC0000}

\begin{abstract}
    We introduce \chigeom{} – a library that generates graph data for testing and benchmarking GNNs' ability to predict chirality. \chigeom{} generates synthetic graph samples with (i) user-specified geometric and topological traits to isolate certain types of samples and (ii) randomized node positions and species to minimize extraneous correlations. Each generated graph contains exactly one chiral center labeled either R or S, while all other nodes are labeled N/A (non-chiral). The generated samples are then combined into a cohesive dataset that can be used to assess a GNN's ability to predict chirality as a node classification task. \chigeom{} allows more interpretable and less confounding benchmarking of GNNs for prediction of chirality in the graph samples which can guide the design of new GNN architectures with improved predictive performance. We illustrate \chigeom{}'s efficacy by using it to generate synthetic datasets for benchmarking various state-of-the-art (SOTA) GNN architectures. The conclusions of these benchmarking results guided our design of two new GNN architectures. The first GNN architecture established all-to-all connections in the graph to accurately predict chirality across all challenging configurations where previously tested SOTA models failed, but at a computational cost (both for training and inference) that grows quadratically with the number of graph nodes. The second GNN architecture avoids all-to-all connections by introducing a virtual node in the original graph structure of the data, which restores the linear scaling of training and inference computational cost with respect to the number of nodes in the graph, while still ensuring competitive accuracy in detecting chirality with respect to SOTA GNN architectures.
\end{abstract}

\section{Introduction}
\label{sec:introduction}

Materials and molecular structures naturally map onto graphs, where atoms are represented as nodes and interatomic bonds as edges. Because of this natural mapping, \acp{gnn} reduce the cumbersome and tedious processes of input feature engineering, which is essential for other types of \ac{dl} models, making them the tool of choice for predicting a wide variety of properties at the atomic scale. By predicting scientifically relevant structural and atomic properties, \acp{gnn} facilitate research and development in chemistry and materials science. As a result, they have important downstream applications in healthcare (drug design), technology (battery materials, semiconductors), sustainability (biodegradable materials), infrastructure (composites, alloys), and agriculture (fertilizers, pesticides).

Various material and molecular properties strongly depend on chirality. Three examples with important practical applications are: (i) binding affinity \citep{gcpnet2022, chienn2023}, (ii) electronic circular dichroism \citep{ecd_former}, and (iii) toxicity \citep{toxicity_prediction}. To accurately predict these properties, it is crucial for \acp{gnn} to correctly capture the chirality of an atomic structure. High-quality benchmarks are critical to guide the development of new, robust \ac{gnn} models because they allow the research community to identify shortcomings of existing \ac{gnn} architectures and propose new \ac{gnn} architectures to overcome them. However, current datasets for benchmarking chirality prediction have serious limitations. Firstly, there is no straightforward way to isolate data samples with geometric and topological traits that are relevant to a {\ac{gnn}}'s predictive ability. Isolating relevant traits, such as the hop distance between the nodes that determine a structure's chirality, enables targeted testing and thereby more interpretable results for analyses of \ac{gnn} prediction failures. Secondly, current benchmark datasets contain extraneous correlations that \acp{gnn} may inadvertently use when making predictions. This may produce misleading results where a \ac{gnn} records high performance on the benchmark without learning the function of chirality. Traits that may have extraneous correlations with the benchmarking data are chemical composition, atomic numbers, and interatomic distances/angles.

To address these limitations, we developed \chigeom{}, a library that generates graphs for testing {\acp{gnn}}' ability to predict chirality. \chigeom{} enables targeted testing by generating datasets composed exclusively of samples with user-specified geometric and topological traits, while minimizing extraneous correlations by randomizing node species and positions. To avoid the injection of extraneous correlations in the data, \chigeom{} generates synthetic graph samples without taking chemical constraints into consideration. As a result, the edges of the generated graph samples do not have chemical meaning and therefore cannot be easily deduced from nodes attributes (e.g., atom types and number of valence electrons for each atom). In contrast, material and molecular graphs typically feature chemically meaningful bonds that are ``learnable'' from atomic properties. This distinction is crucial for benchmarking \ac{gnn} architectures that establish all-to-all connections in the graph (e.g., SE(3)-transformer\footnote{Since there is a mathematical equivalence between the self-attention mechanism in Transformers and all-to-all message passing in GNNs \citep{transformers-are-gnns}, Transformers fall within our broader investigation of GNN architectures.} and global E3NN). In fact, when these models are used on chemical data, they can implicitly reconstruct the original connectivity of the graph samples using the chemically meaningful nodal features. However, such retrievability cannot be performed when the graph samples are synthetically generated by \chigeom{} because the input graph samples do not consider the chemistry. Therefore, \ac{gnn} architectures that establish all-to-all connections in the graph must explicitly keep track of the original graph topology to succeed in benchmarking tests where data is synthetically generated with the \chigeom{} library. 

We illustrate the effectiveness of \chigeom{} by benchmarking the ability of several widely used \ac{sota} \acp{gnn} to model chirality. In doing so, we find that \chigeom{} eliminates a previously observed confounding result (Section~\ref{subsec:benchmarking-existing-architectures}) and yields more interpretable results, which facilitates root-cause-analyses of why \ac{sota} \acp{gnn} fail to predict chirality (Section~\ref{sec:benchmarking}). Based on the shortcomings identified, we propose two new \ac{gnn} architectures. Both are equivariant to reflections \footnote{Although invariance is formally a special case of equivariance (see Appendix~\ref{app:equivariance}), we use ``equivariant'' to concisely refer to models that are equivariant but not invariant.}. The first \ac{gnn} architecture proposed uses global (all-to-all) connections and engineered edge features that preserve the original graph topology, which attains high accuracy across challenging configurations where previous models failed. Its ability to learn long-range interactions while including global connections is reasonably expected, since global connections artificially convert all long-range interactions into direct, short-range ones. However, its computational cost scales quadratically with the number of nodes, which is undesirable for large graph structures. To mitigate the computational cost while preserving the capability to model long-range interactions, inspired by existing work \citep{virtual-node-phonon, virtual-node-protein-binding, virtual-node-large-graphs, virtual-node-hamiltonian}, we introduce a second \ac{gnn} architecture that replaces global connections with the inclusion of a virtual node in the original graph, which is then connected to all the other nodes to accelerate exchange of information across the graph topology. While incorporating a virtual node partially sacrifices accuracy compared to the globally connected model, it maintains linear computational scaling. Notably, it is the only \ac{gnn} we test that is able to detect long-range chirality with better-than-random accuracy while scaling linearly. To the best of our knowledge, this is the first study where the technique of virtual graph nodes is used for chirality predictions. Our numerical results suggest a path towards developing accurate and computationally efficient \ac{gnn} models that capture global chirality, which presents an intriguing direction for future research. The identification of such a class of \ac{gnn} models with judicious balance between training cost and accuracy is relevant for real-world computational chemistry applications. 

The remainder of the paper is organized as follows. In Section~\ref{sec:background}, we provide background on chirality and chirality prediction with \acp{gnn}. In Section~\ref{sec:chi-geometry}, we explain the configurational options that dictate dataset generation in \chigeom{}. In Section~\ref{sec:benchmarking}, we benchmark various \ac{sota} \ac{gnn} architectures, taking advantage of \chigeom{}'s targeted testing and improved interpretability to pinpoint each architecture's shortcomings. Using these insights, we then develop two new \ac{gnn} architectures. The first architecture predicts chirality with high accuracy across all challenging configurations where the other \ac{sota} models failed, but scales quadratically with the number of nodes, while the second architecture strikes a balance between computational efficiency and accuracy and provides a promising direction for future research. Lastly, in Section~\ref{sec:conclusions}, we provide concluding remarks.

\section{Background}
\label{sec:background}

\subsection{Chirality}
\label{subsec:background-chirality}

An object \( \mathrm{O} \) is chiral if there exists no combination of rotations and translations that maps \( \mathrm{O} \) onto its mirror image \( \mathrm{O'} \). As a result, there is a 1-to-1 correspondence between chirality and reflection-orientation. Mathematically, this is expressed as:
\[
\forall\ \text{transformations}\ \mathbb{T} \in \text{SE}(3), \mathbb{T(O)} \neq \mathrm{O'}
\]

\begin{figure}[h!]
    \begin{subfigure}[t]{0.48\textwidth}
        \centering
        \includegraphics[width=0.7\textwidth]{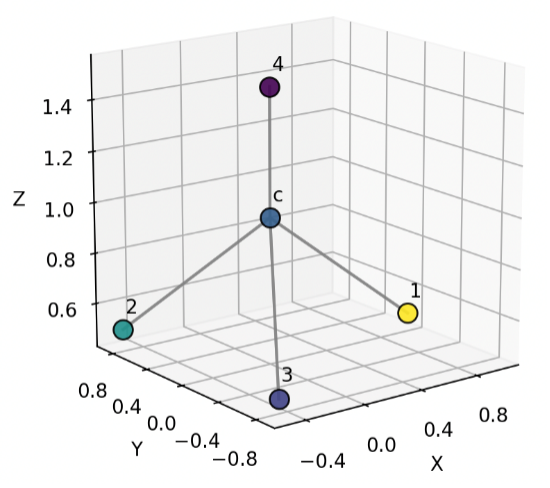}
        \caption{Chiral center has label \textbf{R} (clockwise)}
    \end{subfigure}
    \hfill
    \begin{subfigure}[t]{0.48\textwidth}
        \centering
        \includegraphics[width=0.7\textwidth]{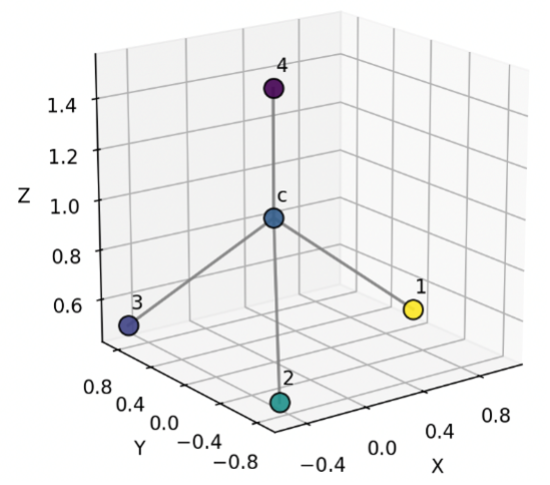}
        \caption{Chiral center has label \textbf{S} (counterclockwise)}
    \end{subfigure}
    \caption{Example of mirrored chiral structures. The chiral center is labeled `c', and its neighbors are labeled numerically according to their priority.}
    \label{fig:chiral-centers-R/S}
\end{figure}

In chemical contexts, the \ac{cip} priority rules are a common procedure for identifying chiral centers and determining their chirality tags \citep{chemistry_nomenclature}. To illustrate, the process goes as follows:
\begin{enumerate}
    \item Identify Chiral Centers: An atom (node) with four unique neighbors is considered a potential chiral center (labeled `c' in Figure~\ref{fig:chiral-centers-R/S}). Otherwise, the atom is labeled N/A.
    \item Assign Priorities: Rank the four neighbors from highest to lowest by atomic number. If two or more neighbors share the same atomic number, continue by recursively comparing node attributes, edge attributes, etc, until all neighbors are unequivocally differentiated (stop and label the node N/A if the neighbors cannot be differentiated).
    \item Orient the Atomic Structure: Position the atomic structure so that the neighbor with the lowest priority points away from the viewer, behind the chiral center.
    \item Determine Chirality: Trace the path along the 1st, 2nd, and 3rd highest-priority neighbors. If the path from \( 1 \to 2 \to 3 \) is clockwise, label the center as R; if it is counterclockwise, label it as S (label N/A if there is no clockwise or counterclockwise orientation) (see Figure~\ref{fig:chiral-centers-R/S} for examples)
\end{enumerate}

\subsection{GNNs for Chirality Prediction}
\label{subsec:background-gnns}
Convolutional \acp{gnn} \citep{gnns} (a.k.a. message-passing \acp{gnn}), in particular, have been shown to effectively model the complex relational structures of graphs via message-passing layers \citep{mfc, message-passing-gnns, schnet, sage, gat, gemnet, gnns-for-atomic-structures, hydragnn, nequip2022, mace2022, m3gnet, liao_equiformer, liao_equiformerv2}. Most \ac{gnn} architectures cannot to predict chirality on graphs, because their representation of different chiral forms is identical. However, some architectures have been developed to include chirality prediction as a capability\citep{gcpnet2022, chienn2023, chiro_2021, spherenet}. Specifically, \acp{gnn} have been applied to predict properties that depend critically on molecular chirality. An example is the binding affinity \citep{gcpnet2022, chienn2023}, which measures the strength of a molecule's bond to a biological target (e.g., small organic molecule binding to a large protein structure). Another example is toxicity \citep{toxicity_prediction}, where the chiral form a molecule takes may drastically change both its intended and/or side effects when used as a drug. A final example is \ac{ecd} \citep{ecd_former}, which refers to the differential absorption of left and right polarized light. \ac{ecd} is highly relevant to the discovery and development of new optical materials. However, many stereochemical assignments rely on long-range structural relationships that \acp{gnn} struggle to accurately capture \citep{vanishing-grad, exploding-grad, oversquashing_observed_and_theory, oversmoothing_observed}. For example, the \ac{cip} procedure for labeling a stereocenter (Section~\ref{subsec:background-chirality}) may depend on nodes that are multiple connection hops away from the chiral center if the immediate neighbors of a chiral center have equivalent traits (e.g. atomic number). Another example arises in axially chiral systems, such as allenes and cumulenes, where chirality is determined by the relative angular orientation of substituents separated by multiple double bonds (for example, 1,3-diphenylallene). \citet{matrixfunction2024} specifically employed cumulenes to show that accurately predicting force-field energies requires \acp{gnn} capable of modeling these long-range interactions. But, message-passing \ac{gnn} architectures with K message-passing layers cannot propagate information beyond K hops \citep{oversmoothing_observed_and_theory}. Moreover, although arbitrarily deep networks can theoretically propagate information from arbitrarily many hops away, they encounter several computational challenges. First, over-smoothing causes node embeddings to converge toward indistinguishable vectors, degrading their utility for downstream tasks \citep{oversmoothing_observed}. Second, deep \acp{gnn} are prone to vanishing or exploding gradients, which hinder effective training \citep{vanishing-grad, exploding-grad}. Finally, over-squashing arises when exponentially larger neighborhoods are compressed into fixed-size representations, attenuating the very long-range signals that deeper layers aim to capture \citep{oversquashing_observed_and_theory}. 

Since long-range dependencies influence chirality assignments and are poorly captured by standard \acp{gnn}, it is important to construct test sets that emphasize these interactions when evaluating model robustness. However, existing benchmarks neither permit the selective sampling of molecules according to specific geometric or topological criteria nor avoid spurious correlations that can obscure a model's true capabilities. For example, \citet{gcpnet2022} found that the DimeNet++ architecture was able to classify a chiral molecule's chirality as R or S with 65.7\% accuracy. This result is surprising because the architecture in question uses exclusively the atomic number, pairwise distances, and triplet angles as inputs, all of which do not change under reflection. Consequently, the representations are identical for two mirror-image atomic structures, which should imply near-random (i.e., 50\%) accuracy for R/S chirality classification. Yet, its accuracy significantly exceeded random chance, which suggests that the model used extraneous correlations to predict chirality. The lack of targeted datasets that minimize extraneous correlations, which lead to confounding results like the one mentioned above, is the motivation behind developing \chigeom{}, a library for generating randomized chiral graphs with explicit control over relevant geometric/topological features.

\section{\chigeom{}}
\label{sec:chi-geometry}

The dataset generation of \chigeom{} is parametrized by the following arguments: \textbf{Chirality Distance}, \textbf{Chirality Type}, \textbf{Species Range}, and \textbf{Noise}. The Chirality Distance and Chirality Type parameters control the topological/geometric features of the generated dataset, as illustrated by Figure~\ref{fig:chigeo-table} and explained in Appendix~\ref{app:data-generation}. Additionally, \chigeom{} reduces the effect of potential extraneous correlations by randomly assigning the node species (always) and positions (when Noise = True). Lastly, the species range varies the degree of difficulty of the preliminary sorting task that is required to predict chirality, which is useful for faster testing and development. In each structure, there is exactly one node that is a chiral center with label R or S (see details in Appendix~\ref{app:data-generation}). Below, we briefly elaborate on how the configurational parameters dictate sample generation.

\begin{figure}[ht]
    \centering
    \includegraphics[width=0.6\linewidth]{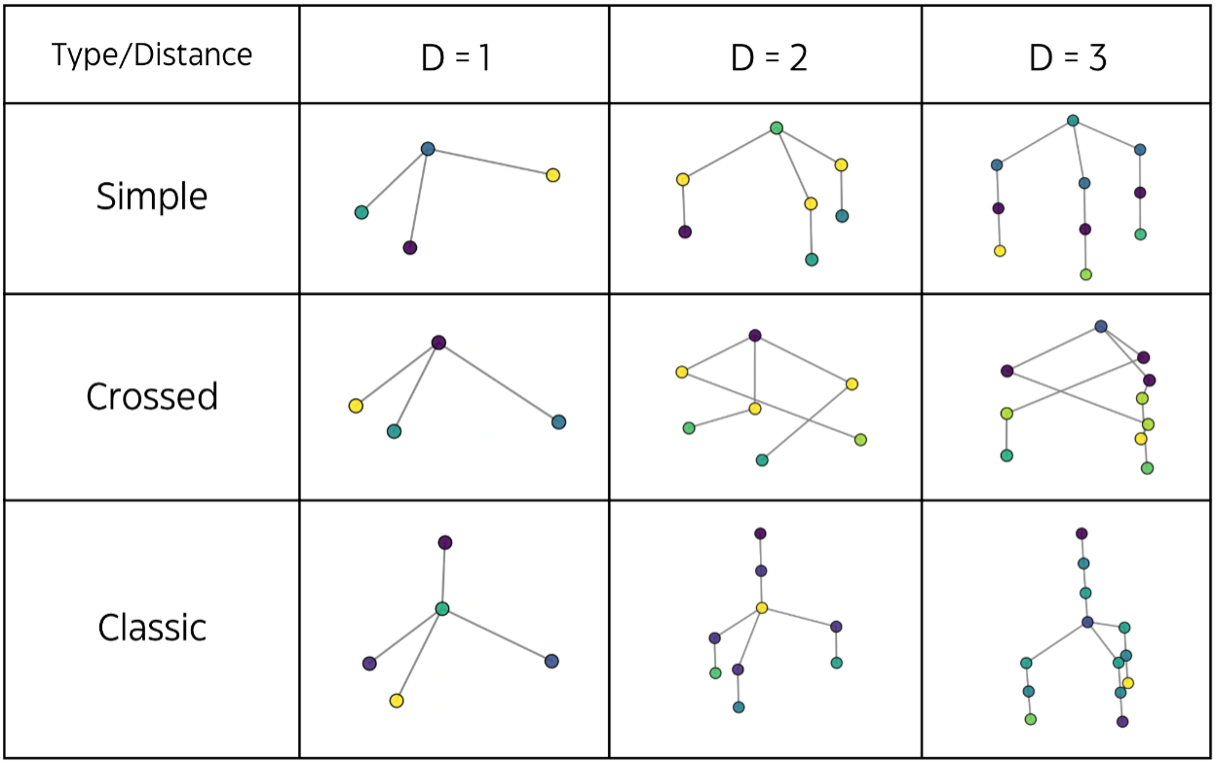}
    \caption{Examples of the Simple, Crossed, and Classic configuration types in \chigeom{} at distances 1, 2, and 3 with species range 15 and noise = False.}
    \label{fig:chigeo-table}
\end{figure}

\textbf{Chirality Distance:} Specifies how many connection `hops' the chiral relationship spans by adding (d-1) intermediate `layers'\footnote{The `layers' of samples are unrelated to the convolutional layers of graph neural networks. Refer to Figure~\ref{fig:chigeo-table} for an illustration of increasing layers as the chirality distance increases, or Appendix~\ref{app:data-generation} for an explanation of layer generation.} in between the chiral center and the nodes that determine its chirality, which is crucial for testing the robustness of the {\ac{gnn}}'s ability to learn long-range interactions that determine the chirality.

\textbf{Chirality Type:} Provides three choices for structural arrangements of chiral configurations: Simple, Crossed, and Classic, each testing different aspects of the {\ac{gnn}}'s ability to learn chirality. The classic and simple types test the {\ac{gnn}}'s ability to learn from variable-range node attributes and local geometry. On the other hand, the crossed type tests the {\ac{gnn}}'s ability to learn from variable-range node attributes and geometry (Appendix~\ref{subapp:sample-types-test-for}). Since chirality is determined by both node attributes and geometry, we constructed the sample types to span a comprehensive set of attribute-geometry combinations. As a result, the available sample types capture nearly all structural configurations of interest for testing chirality prediction. Additionally, the classic type more closely represents the types of chiral centers found in chemistry (four neighbors), while the simple type is the minimal configuration needed to fit the formal definition of chirality (three neighbors). To provide a more complete set of options to users with different needs and preferences, we included both.

\textbf{Species Range:} For each generated chiral structure, the integer node species are selected randomly and uniformly from the set \{1, 2, …, species range\}. Some nodes may be constrained to share the same node species, but all the selected node species are chosen at random. So, the species range parameter sets the upper bound for the integer selection. As explained in (Section~\ref{subsec:background-chirality}) and Appendix~\ref{app:data-generation}, labeling a chiral center requires ranking nodes based on their attributes. By varying the species range parameter, users can adjust the complexity of this ranking task that the \ac{gnn} must learn. By setting a reasonably small species range, users can test chirality prediction for \acp{gnn} without requiring a large dataset for the \ac{gnn} to learn large-scale sorting. This is useful for running quicker tests during development.

\textbf{Noise:} If noise = True, the positions of the nodes will be randomly assigned. This randomization is useful to reduce the effect of extraneous correlations, while noise = False is clearer for visualization.

\section{Benchmarking Results}
\label{sec:benchmarking}

By specifying \chigeom{}'s chirality distance parameter, we systematically tested several widely used \ac{gnn} architectures on their ability to identify chirality at specified hop distances (see Appendix~\ref{app:hyperparameters} for more details on the dataset, model, and training parameters). For each \ac{gnn} architecture, we evaluated performance on homogeneous datasets where chirality is consistently determined by nodes that are all at hop distance D from the chiral center. First, we train and evaluate an architecture on a dataset with D = 1. If a model does not achieve high accuracy, we repeat the experiment nine more times (10 runs total) to verify the result and produce error bars. If a model succeeds at D = 1, we extend evaluation to datasets with hop distances D = 2 - 9 and do not repeat the runs because of the computational cost. All models used the same datasets for each respective D value. The homogeneity of the datasets in our tests yields more interpretable results, which we used to develop two new architectures that perform better across the challenging configurations where the other models failed. Specifically, we tested five existing architectures\textemdash{} DimeNet++ \citep{dimenet++}, Vanilla MPNN, SE(3)-Transformer \citep{lucidrains_equiformer}, E3NN\footnote{The e3nn library (MIT license) \citep{e3nn} is an open-source software for building equivariant neural networks. For our analysis, we use the January 2021 release of its example model.} with global connections \citep{e3nn}, and E3NN with local connections \citep{e3nn}\textemdash{} as well two developed architectures\textemdash{}E3NN with global connections and feature engineering \citep{e3nn} and E3NN with virtual node \citep{e3nn}.

The tested models include those that are equivariant to reflections (all E3NN models), those that are invariant to reflections (DimeNet++, SE(3)-Transformer) (Appendix~\ref{app:equivariance}), and those that are neither invariant nor equivariant (Vanilla MPNN). Additionally, the tested models include various connectivity types (global, original, virtual node). Thus, our selection of tested models collectively spans a broad spectrum of architectural designs relevant to chirality prediction.

The number of message‐passing layers was chosen to ensure sufficient receptive field given each model’s connectivity and the value of D. For models using the original adjacency (DimeNet++, Vanilla MPNN, Local E3NN), we set layers = D + 3 because preliminary experiments showed four layers sufficed to predict chirality at D = 1 with the Local E3NN architecture, and we infer that the number of required layers will increase correspondingly with D. For architectures that use all-to-all connections (Global E3NN, E3NN with Virtual Node, SE(3)-Transformer) and thereby reduce the hop distance between nodes in the graph, we use four layers.

\subsection{Existing Architectures}
\label{subsec:benchmarking-existing-architectures}
We examine five existing architectures on their ability to classify chirality, which are listed below:

\begin{table*}[ht]
    \centering
    \begin{tabularx}{\textwidth}{p{0.15\textwidth} X}
    \toprule
    \textbf{Model} & \textbf{Architectural Details} \\
    \midrule
    DimeNet++ \citep{dimenet++} & 
    Includes pairwise distances and triplet angles in message-passing. \\
    \specialrule{0pt}{0pt}{4pt}
    
    Vanilla MPNN & 
    Relies on a pairwise message-passing layer without equivariance to reflections, which is shared across many models (e.g., GCNN \citep{message-passing-gnns}, SAGE \citep{sage}, MFC \citep{mfc}, GATv2 \citep{gat-v2}). Node positions are appended directly to the node features to ensure that the model is theoretically capable of modeling chirality. \\
    \specialrule{0pt}{0pt}{4pt}
    
    SE(3)-Transformer \citep{lucidrains_equiformer}\footnotemark & 
    Equivariant to rotations and translations and has shown strong performance on structural and atomic prediction tasks \citep{liao_equiformer, fuchs_se3transformer, omat24}. \\
    \specialrule{0pt}{0pt}{4pt}
    
    \parbox[t]{\linewidth}{E3NN \\ (two variants: Local, Global) \citep{e3nn}} &
    The e3nn library has been used to develop several notable \acp{gnn} \citep{nequip2022, mace2022}. We use the e3nn library to construct \ac{gnn} architectures that are equivariant to reflections, and thereby are able to detect chirality (Appendix~\ref{app:equivariance}). We evaluate two variants: the local model which uses the original graph adjacency, and the global model which rewires the graph for all-to-all connections. \\

    \bottomrule
    \end{tabularx}
    \caption{The existing architectures which are tested with \chigeom{} and their details.}
    \label{tab:existing-architectures}
    \vspace{-3mm}
\end{table*}

\footnotetext[5]{The open-source PyTorch Implementation of the SE(3)-Transformer (a.k.a. Equiformer or Equivariant Transformer) \citep{lucidrains_equiformer} draws on previous publications from \citet{lucid1, lucid2, lucid3, lucid4, lucid5, lucid6, lucid7, lucid8, lucid9}.}

\begin{table}[h!]
\vspace{-0mm}
\centering
\renewcommand{\arraystretch}{1.1}
\scriptsize
\begin{tabular}{|l|c|c|c|c|}
\hline
\rowcolor{gray!40}
\textbf{Class} & \textbf{DimeNet++ (D=1)} & \textbf{Vanilla MPNN (D=1)} & \textbf{SE(3)-Transformer (D=1)} & \textbf{Global E3NN (D=1)} \\
\hline
N/A & \cellcolor{my100!75}100\% \( \pm \) 0\% & \cellcolor{my100!75}100\% \( \pm \) 0\% & \cellcolor{my79!75}79\% \( \pm \) 1\% & \cellcolor{my76!75}76\% \( \pm \) 2\% \\
\hline
R   & \cellcolor{my51!75}51\% \( \pm \) 18\% & \cellcolor{my58!75}58\% \( \pm \) 9\% & \cellcolor{my38!75}38\% \( \pm \) 19\% & \cellcolor{my65!75}65\%  \( \pm \) 2\% \\
\hline
S   & \cellcolor{my49!75}49\% \( \pm \) 18\% & \cellcolor{my42!75}42\% \( \pm \) 9\% & \cellcolor{my41!75}41\% \( \pm \) 19\% & \cellcolor{my66!75}66\% \( \pm \) 2\% \\
\hline
\end{tabular}
\caption{Classification accuracy of the DimeNet++, Vanilla MPNN, SE(3)-Transformer, and Global E3NN architectures at hop distance D = 1 over 10 repetitions with standard deviation error bars.}
\label{tab:combined_model_accuracy}
\vspace{-6mm}
\end{table}

As shown in Table~\ref{tab:combined_model_accuracy}, the DimeNet++, Vanilla MPNN, SE(3)-Transformer, and Global E3NN architectures failed to predict chirality with high accuracy at D = 1. Both the DimeNet++ and SE(3)-Transformer models distinguished R or S chiral centers at no better than random, as expected given their reflection invariance (Appendix~\ref{app:equivariance}), which precludes any representation of chirality. 

While the Vanilla MPNN is not reflection-invariant and therefore, in principle, should be capable of predicting chirality, it performed no better than random when classifying chiral centers as R or S. This highlights practical limitations of GNN architectures in modeling chirality, despite their theoretical capability to do so. Both the SE(3)-Transformer and Global E3NN models struggled to classify non-chiral centers as N/A, which we attribute to losing clear topological indicators for which node is a chiral center when rewiring the graph with global connections. Again, we emphasize that in real-world molecular and materials datasets typically have chemically meaningful bonds that the \ac{gnn} may be able to learn. However, this is not true for graphs generated by \chigeom{} because they are randomized to reduce extraneous correlations. In predictive tasks where the edges are learnable by the \ac{gnn}, the SE(3)-Transformer and E3NN model with global connections may perform better, although we expect the SE(3)-Transformer to always predict chiral centers as R or S no better than random because of its reflection invariance (Appendix~\ref{app:equivariance}).

These results highlight two key contributions from \chigeom{}. First, it eliminates the confounding result reported by \citep{gcpnet2022}, where DimeNet++ classified chirality at above random accuracy despite being reflection-invariant (Section~\ref{subsec:background-gnns}). Second, it shows that the Vanilla MPNN fails to learn chirality even in the simplest setting (D = 1) despite architecture's theoretical capacity to model chirality, which was not obvious a-priori.

\begin{table}[h!]
\vspace{-3mm}
\centering
\renewcommand{\arraystretch}{1.0}
\scalebox{1.0}{
\begin{tabular}{|>{\scriptsize}l|>{\scriptsize}c|>{\scriptsize}c|>{\scriptsize}c|>{\scriptsize}c|
                >{\scriptsize}c|>{\scriptsize}c|>{\scriptsize}c|>{\scriptsize}c|>{\scriptsize}c|}
\hline
\rowcolor{gray!40}
\textbf{Class} & 
\textbf{D=1} & 
\textbf{D=2} & 
\textbf{D=3} & 
\textbf{D=4} & 
\textbf{D=5} & 
\textbf{D=6} & 
\textbf{D=7} & 
\textbf{D=8} & 
\textbf{D=9} \\
\hline
N/A & 
\cellcolor{my100!70} 100\% & 
\cellcolor{my94!70}  94\% & 
\cellcolor{my97!70}  97\% & 
\cellcolor{my92!70}  92\% & 
\cellcolor{my92!70}  92\% &
\cellcolor{my56!70}  56\% &
\cellcolor{my26!70}  26\% &
\cellcolor{my44!70}  44\% &
\cellcolor{my39!70}  39\% \\
\hline
R & 
\cellcolor{my95!70}  95\% & 
\cellcolor{my65!70}  65\% & 
\cellcolor{my77!70}  77\% & 
\cellcolor{my68!70}  68\% & 
\cellcolor{my69!70}  69\% &
\cellcolor{my16!70}  16\% &
\cellcolor{my64!70}  64\% &
\cellcolor{my33!70}  33\% &
\cellcolor{my38!70}  38\% \\
\hline
S & 
\cellcolor{my99!70}  99\% &
\cellcolor{my33!70}  33\% &
\cellcolor{my21!70}  21\% &
\cellcolor{my31!70}  31\% &
\cellcolor{my29!70}  29\% &
\cellcolor{my20!70}  20\% &
\cellcolor{my15!70}  15\% &
\cellcolor{my20!70}  20\% &
\cellcolor{my29!70}  29\% \\
\hline
\end{tabular}
}
\caption{Classification accuracy of the Local E3NN architecture at hop distances D = 1-9.}
\label{tab:e3nn_local}
\vspace{-6mm}
\end{table}

\begin{figure}[h!]
\vspace{-3mm}
    \centering
    \begin{subfigure}[t]{0.48\textwidth}
        \centering
        \includegraphics[width=0.75\textwidth]{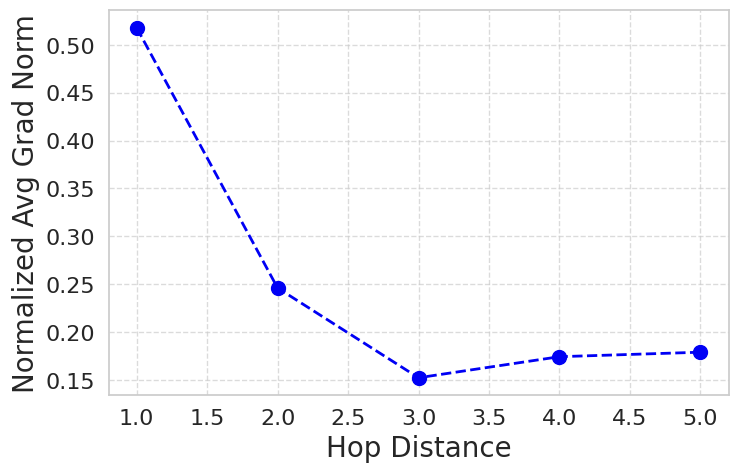}
        \vspace{-2mm}
        \caption{Normalized average gradient of node features that determine the chiral center's tag, as a function of hop distance.}
        \label{fig:gradient_signal}
    \end{subfigure}
    \hfill
    \begin{subfigure}[t]{0.48\textwidth}
        \centering
        \includegraphics[width=0.75\textwidth]{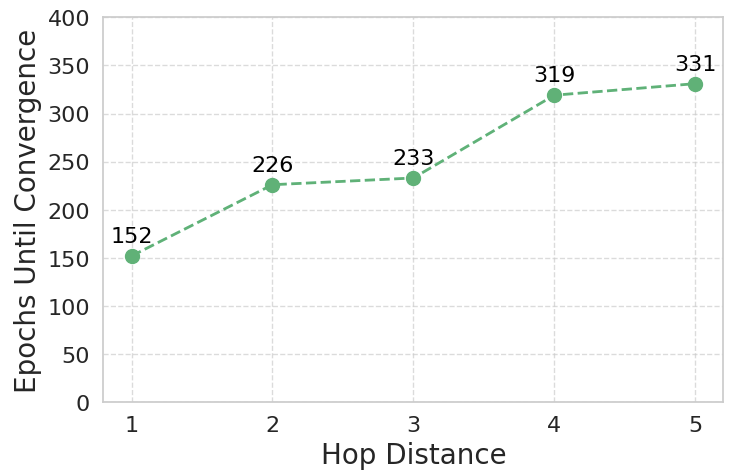}
        \vspace{-2mm}
        \caption{Epochs required for convergence by hop distance.}
        \label{fig:epochs_required}
    \end{subfigure}
    \vspace{-2mm}
    \caption{Indicators of over-squashing and extended training times as hop distance increases.}
    \label{fig:local_e3nn_analysis}
\vspace{-3mm}
\end{figure}

The Local E3NN model achieved high prediction accuracy for D = 1 but degrades rapidly for higher distances Table~\ref{tab:e3nn_local}. This is in line with well-established limitations of deep \acp{gnn}, namely their difficulty in capturing long-range dependencies \citep{oversmoothing_observed, oversmoothing_observed_and_theory, oversquashing_theory, oversquashing_observed_and_theory} and vanishing and/or exploding gradients \citep{vanishing-grad, exploding-grad}. At D \( \geq \) 6, the model encounters training instabilities including returning losses of inf and gradients of NaN, despite gradient clipping in the training stage. We also found strong indications of over-squashing \citep{oversquashing_theory, oversquashing_observed_and_theory} via the normalized average gradient of node features with respect to the chiral center's classification loss (Appendix~\ref{app:oversquashing}). As seen in Figure~\ref{fig:gradient_signal}, the gradients of the node features which determine a chiral center's tag diminished as their hop distance from the chiral center increased, despite the importance of these features for the classification. Additionally, the training took more epochs to converge as D increased (Figure~\ref{fig:epochs_required}), suggesting that there are more instabilities and/or weaker signals that delay convergence for larger hop distances.

Overall, we see indications that the Local E3NN model architecture struggles to learn chirality at hop distances greater than 1 due to over-squashing, weak training signals, and instability.

\subsection{Developed Architectures}
\label{subsec:benchmarking-developed-architectures}
Based on these results, we developed two new architectures designed to address all the limitations discussed in Section~\ref{subsec:benchmarking-existing-architectures}. Namely, the shortcomings are that: (1) DimeNet++, the General MPNN, and the SE(3)-Transformer do not detect reflections, (2) the E3NN model with local connections exhibited over-squashing and training instabilities, and (3) the E3NN model with global connections and the SE(3)-Transformer lose topological information during graph rewiring. 

The first model we developed is an E3NN model with global connections and judiciously engineered edge features that preserve the original graph topology. The second model is an E3NN model with a virtual node. Both models are equivariant to reflections (and therefore chirality) by construction (Appendix~\ref{app:equivariance}). The global connections mitigate over-squashing \citep{oversquashing_observed_and_theory, oversquashing_theory} by providing direct, all-to-all connections that avoid long-range bottlenecks. Although all long-range information still must pass through a single virtual node (creating a bottleneck), using a virtual node reduces the number of edges that information must propagate through to reach distant nodes, thereby mitigating over-squashing. Also, both models use only four message-passing layers, thereby minimizing vanishing and/or exploding gradients which often arise in deep networks \citep{vanish-exploding-grad-GNNs, exploding-grad}.

\subsubsection{E3NN with Global Connections and Feature Engineering}
\label{subsubsec:benchmarking-e3nn-global-fe}
Firstly, we developed an E3NN model with global connections that was adjusted to include judiciously engineered edge features which preserve the original graph topology. Specifically, we use (1) the shortest-path hop distance under the original adjacency, and (2) a one-hot encoding denoting whether that edge existed in the original graph. By preserving the original graph topology via the edge features, the network learns to disambiguate the chiral center even after rewiring with global connections.

\begin{table}[h!]
\vspace{-3mm}
\centering
\renewcommand{\arraystretch}{1.0}
\scalebox{1.0}{
\begin{tabular}{|>{\scriptsize}l|>{\scriptsize}c|>{\scriptsize}c|>{\scriptsize}c|>{\scriptsize}c|>{\scriptsize}c|
                 >{\scriptsize}c|>{\scriptsize}c|>{\scriptsize}c|>{\scriptsize}c|}
\hline
\rowcolor{gray!40}
\textbf{Class} & 
\textbf{D=1} & 
\textbf{D=2} & 
\textbf{D=3} & 
\textbf{D=4} & 
\textbf{D=5} & 
\textbf{D=6} & 
\textbf{D=7} & 
\textbf{D=8} & 
\textbf{D=9} \\
\hline
N/A & 
\cellcolor{my100!70} 100\% &    
\cellcolor{my100!70} 100\% &    
\cellcolor{my100!70} 100\% &    
\cellcolor{my100!70} 100\% &    
\cellcolor{my100!70} 100\% &    
\cellcolor{my100!70} 100\% &    
\cellcolor{my99!70}  99\%  &    
\cellcolor{my100!70} 100\% &    
\cellcolor{my100!70} 100\% \\   
\hline
R & 
\cellcolor{my100!70} 100\% &    
\cellcolor{my98!70}  98\%  &    
\cellcolor{my100!70} 100\% &    
\cellcolor{my98!70}  98\%  &    
\cellcolor{my100!70} 100\% &    
\cellcolor{my99!70}  99\%  &    
\cellcolor{my98!70}  98\%  &    
\cellcolor{my100!70} 100\% &    
\cellcolor{my100!70} 100\% \\   
\hline
S & 
\cellcolor{my100!70} 100\% &    
\cellcolor{my98!70}  98\%  &    
\cellcolor{my100!70} 100\% &    
\cellcolor{my97!70}  97\%  &    
\cellcolor{my100!70} 100\% &    
\cellcolor{my100!70} 100\% &    
\cellcolor{my98!70}  98\%  &    
\cellcolor{my99!70}  99\%  &    
\cellcolor{my99!70}  99\%  \\   
\hline
\end{tabular}
}
\caption{Classification accuracy of an E3NN model with global connections and feature engineering.}
\label{tab:e3nn_global_fe}
\vspace{-6mm}
\end{table}

As shown in Table~\ref{tab:e3nn_global_fe}, the E3NN model with this combination of global connections and judiciously designed edge features predicts chirality with high accuracy on all configuration types that we previously used to test the other models. However, it is important to highlight that the model's successful capture of long-range interactions is not particularly surprising given its global (all-to-all) connectivity, which scales quadratically with the number of nodes and naturally facilitates information flow across any hop distance. Additionally, the engineered features preserving the original graph topology were necessary specifically because \chigeom{} generates synthetic graphs where edges cannot be inferred from node attributes alone. Nevertheless, this model's performance confirms \chigeom{}'s capability as a practical benchmark: it identifies whether models can effectively learn to predict chirality on graph-structured data, especially in difficult samples with long-range interactions, thereby guiding future model improvements.

\subsubsection{E3NN with Virtual Node}
\label{subsubsec:benchmarking-e3nn-virtual}
Second, we developed an E3NN model with a virtual node to attempt to capture long-range interactions that determine the chirality while scaling linearly with the number of nodes. A virtual node is an additional node, not part of the original graph, that receives messages from all other nodes (all-to-one) and sends messages back to them (one-to-all). Message-passing with a virtual node scales linearly with the number of nodes in a graph. Additionally, previous studies have shown that, in principle, \acp{mpnn} with virtual node can, arbitrarily approximate the self-attention layer of an all-to-all message passing model (Graph Transformer). Given (i) the success of the all-to-all message-passing model in Section~\ref{subsubsec:benchmarking-e3nn-global-fe}, (ii) the previous work establishing the ability for virtual nodes to approximate all-to-all message passing, and (iii) the favorable computational scaling of a virtual node as opposed to all-to-all message-passing, we view it as a natural next step to study the chiralty prediction capability of an E3NN model with a virtual node. Prior work has used virtual nodes in other settings than chirality prediction \citep{virtual-node-phonon, virtual-node-protein-binding, virtual-node-large-graphs, virtual-node-hamiltonian}. However, to our knowledge, this is the first study that (i) uses a virtual node which is equivariant to reflections and (ii) applies a virtual node architecture to chirality prediction.

\begin{table}[h!]
\vspace{-3mm}
\centering
\renewcommand{\arraystretch}{1.0}
\scalebox{1.0}{
\begin{tabular}{|>{\scriptsize}l|>{\scriptsize}c|>{\scriptsize}c|>{\scriptsize}c|>{\scriptsize}c|>{\scriptsize}c|
                 >{\scriptsize}c|>{\scriptsize}c|>{\scriptsize}c|>{\scriptsize}c|}
\hline
\rowcolor{gray!40}
\textbf{Class} & 
\textbf{D=1} & 
\textbf{D=2} & 
\textbf{D=3} & 
\textbf{D=4} & 
\textbf{D=5} & 
\textbf{D=6} & 
\textbf{D=7} & 
\textbf{D=8} & 
\textbf{D=9} \\
\hline
N/A & 
\cellcolor{my84!70}  84\% &     
\cellcolor{my87!70}  87\% &     
\cellcolor{my77!70}  77\% &     
\cellcolor{my75!70}  75\% &     
\cellcolor{my63!70}  63\% &     
\cellcolor{my68!70}  68\% &     
\cellcolor{my56!70}  56\% &     
\cellcolor{my85!70}  85\% &     
\cellcolor{my66!70}  66\% \\    
\hline
R & 
\cellcolor{my99!70}  99\% &     
\cellcolor{my78!70}  78\% &     
\cellcolor{my61!70}  61\% &     
\cellcolor{my55!70}  55\% &     
\cellcolor{my55!70}  55\% &     
\cellcolor{my42!70}  42\% &     
\cellcolor{my57!70}  57\% &     
\cellcolor{my45!70}  45\% &     
\cellcolor{my62!70}  62\% \\    
\hline
S & 
\cellcolor{my92!70}  92\% &     
\cellcolor{my57!70}  57\% &     
\cellcolor{my40!70}  40\% &     
\cellcolor{my42!70}  42\% &     
\cellcolor{my43!70}  43\% &     
\cellcolor{my54!70}  54\% &     
\cellcolor{my41!70}  41\% &     
\cellcolor{my45!70}  45\% &     
\cellcolor{my39!70}  39\%  \\   
\hline
\end{tabular}
}
\caption{Classification accuracy of an E3NN model with a virtual node.}
\label{tab:e3nn_virtual_node}
\vspace{-6mm}
\end{table}

The E3NN model with a virtual node yields promising results, achieving better-than-random classification of R/S chiral centers at D = 2 Table~\ref{tab:e3nn_virtual_node}. Notably, no other model that we tested achieves better than random R/S accuracy at D = 2 while computationally scaling linearly with the number of nodes. These results suggest that the virtual node architecture offers a promising trade-off: while not as accurate as the global model, it provides improved performance over other linearly-scaling models at intermediate distances with significantly reduced computational cost.

\begin{figure}[h!]
\vspace{-3mm}
    \centering
    \begin{subfigure}[t]{0.35\textwidth}
        \includegraphics[width=\textwidth]{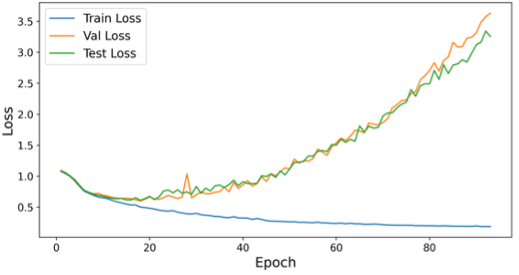}
        \vspace{-5mm}
        \caption{Train/val/test loss over epochs for D = 2.}
        \label{fig:e3nn-virtual-loss}
    \end{subfigure}
    \hfill
    \begin{subfigure}[t]{0.60\textwidth}
        \includegraphics[width=\textwidth]{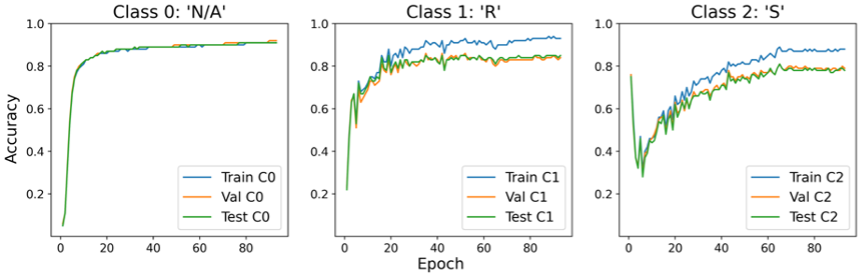}
        \vspace{-5mm}
        \caption{Classification accuracy by class for train/val/test over epochs at D = 2.}
        \label{fig:e3nn-virtual-accuracy}
    \end{subfigure}
    \label{fig:e3nn-virtual-training}
\vspace{-3mm}
\end{figure}

The training dynamics for D = 2 shown in Figures~\ref{fig:e3nn-virtual-loss} and ~\ref{fig:e3nn-virtual-accuracy} help explain these failures and offer avenues for future study. The model reaches its minimum validation loss at epoch 18, but both validation and test accuracies continue rising beyond that epoch. We ascribe this counterintuitive behavior to the model confidently misclassifying a minority of the samples, despite improving on most. At the last epoch, the model's train/test accuracies are [0.91,0.93,0.88] and [0.91,0.85,0.78], respectively, which is not captured in Table~\ref{tab:e3nn_virtual_node} because of early-stopping. We observed similar behavior for all D = 1–9. Overall, these results suggest that although the current virtual-node E3NN does not beat random guessing on R/S classification for D > 2, systematic architectural and algorithmic refinements (e.g. entropy regularization) could yield a variant that retains linear scaling with accuracy competitive to the more expensive global models.
\section{Conclusions}
\label{sec:conclusions}
We have presented \chigeom{}, a library that generates datasets for testing and benchmarking chirality prediction of \acp{gnn}. \chigeom{} allows users to select structurally relevant geometric and topological traits and reduces the effect of extraneous correlations in its dataset generation. By allowing users to select structural features of interest for targeted testing, \chigeom{} gives more interpretable results for a root-cause analysis of prediction failures. Additionally, by reducing the effect of extraneous correlations, \chigeom{} minimizes confounding results whereby a model seems to perform well despite not actually learning the function of chirality. Using \chigeom{} datasets, we illustrated the limitations of several widely used \ac{sota} \ac{gnn} models to classify chirality on graphs. Specifically, we identified over-squashing/training instability, lack of chirality representation, and information loss in graph rewiring as root causes of prediction failures. In addition, we used the insights drawn from our targeted tests to develop two new architectures – an E3NN model with global connections and engineered edge features and an E3NN model with a virtual node – which show improved performance on the challenging configurations where other models failed to predict chirality. Overall, \chigeom{} generates targeted datasets for testing and benchmarking, producing more interpretable and less confounding results, thereby facilitating the development of improved \ac{gnn} architectures for chirality prediction.

\bibliographystyle{plainnat}
\bibliography{references}

\appendix
\section{Dataset, Training, and Model Hyperparameters}
\label{app:hyperparameters}

\paragraph{Dataset}
All \chigeom{} datasets used in our experiments contain 25,000 samples of simple type with species range = 15 and noise = True (see Section~\ref{sec:chi-geometry}, Appendix~\ref{app:data-generation} for details on dataset generation). The train/val/test split was set to 80/10/10.

\paragraph{Training Scheme and Model}
Training Scheme and Model: All models were trained using 500 epochs maximum, a 75-epoch early stopping criterion, 25-epoch learning rate decay with decay factor 0.5, and cross-entropy loss for classification. We used a learning rate of 0.0003, the Adam optimizer, gradient clipping with max \( L_2 \)-norm of 10, batch size of 32, and consistent dimensionalities for node features\footnote{Node features are mapped to input dim of 40 before message-passing, hidden dim of 20 during message-passing, and output dim of 10 after messing-passing.}. Additionally, we set the number of message-passing layers according to what we determine to be the minimum number that is sufficient to learn chirality.

\paragraph{Cross-Entropy Loss}
We use the weighted cross-entropy loss, which is a common loss function for machine learning classification tasks. It is defined as:
\[
\text{CE} = -\sum_{t \in \{\text{N/A},R,S\}} w_t\,p(t)\log(q(t))
\]
where \( p(t) \) is the true probability distribution, \( q(t) \) is the model's predicted probability distribution of a sample to belong to class N/A, R, or S, and \( w_t \) is the weight assigned to a certain class.

\paragraph{Class Weights to Counteract Dataset Imbalance}
The class weights for [N/A, R, S] in the cross-entropy loss are adjusted dynamically to counteract the effect of class imbalance in the datasets. For a dataset of simple type and chirality distance = D we know that \( n_{\text{N/A}} = 3\mathrm{D} \times (n_R + n_S) \), where \( n_i \) indicates the number of nodes of class \( i \). The class weights mirror the imbalance by also being proportional to \( (3 \times \mathrm{D}) \). The general formula for class weights is:
\[
[w_{\text{N/A}}, w_R, w_S] = [1.0,\, b + 3\times \mathrm{D},\, b + 3\times \mathrm{D}]
\]
where \( w_i \) represents the weight for class \( i \) and \( b \) is a constant that changes depending on the model architecture\footnote{Preliminary tests indicated that the chosen values of b produced a sufficiently strong R/S classification signal, stable training, and avoided significant class imbalance. Specifically, we set \( b = 10 \) for the DimeNet++, Vanilla MPNN, Local E3NN, Global E3NN with feature engineering, and E3NN with virtual node models, \( b = 6.0 \) for the Global E3NN, and \( b = 0.0 \) for the SE(3)-Transformer.}.

\section{\chigeom{} Data Generation}
\label{app:data-generation}

\subsection{Sample Generation}
\label{subapp:sample-generation}
Below, we elaborate on the creation process for samples generated by \chigeom{}. The creation starts with the chiral center and proceeds layer by layer along the z axis, where the number of layers is the user supplied parameter chirality distance (a.k.a. D) (Section~\ref{sec:chi-geometry}). Throughout this appendix section, we use \( v \sim \mathrm{U(a,b)} \) to indicate that a variable \( v \) is drawn uniformly at random from the interval [a, b].

\subsubsection{Simple Configurations}
\label{subsubapp:simple-configurations}
For the \chigeom{} samples of simple type, each layer 1-D contains a triplet of nodes. The procedure goes as follows:

\textbf{\underline{Setup:}}\\[4pt]
We begin by randomly selecting node species for all nodes in the configuration. Specifically, we sample 4 + (D - 1) node species without replacement from the interval [1, species range]. This accounts for one species for the chiral center, three distinct species for the final layer, and D - 1 species for the D – 1 intermediate layers, where each intermediate layer shares the same species across all three of its nodes. Let S be this set of sampled node species and be zero-indexed (i.e. its indices go from zero to D + 3).
\[
\mathrm{S} \sim \text{UniformSubset}((1, \text{species range}), 4+(\mathrm{D}-1))
\]

\textbf{\underline{Chiral Center:}}\\[4pt]
\underline{Positions:}

\renewcommand{\arraystretch}{1.2}
\begin{tabular}{|c|c|}
\hline
\textbf{Noise Flag} & \( \boldsymbol{p_c} \) \\ 
\hline
noise = False & \( (0,0,1) \) \\
\hline
noise = True & \( (r\cdot\cos\phi,\; r\cdot\sin\phi,\; 1) \) \\
\hline
\end{tabular}\\[2pt]
where \( p_c \) indicates the position of the chiral center, \( r \sim \mathrm{U}(0,1) \), and \( \phi \sim \mathrm{U}(0,2\pi) \).

\underline{Node Species:}\\[4pt]
\( x_c = \mathrm{S}_0 \)

\vspace{6pt}

\underline{\textbf{For Layer in range (1, D):}}\\[4pt]
\underline{Positions:}\\[4pt]
If Layer = 1, initialize \( z = 1 \).

\begin{enumerate}
  \item Choose parameters

  \vspace{4pt}
  \renewcommand{\arraystretch}{1.5}
  \begin{tabular}{|c|c|c|c|}
    \hline
    \textbf{Noise Flag} & \( \boldsymbol{\Delta z} \) & \( \boldsymbol{r} \) & \( \boldsymbol{\theta} \)\\
    \hline
    noise = False & 0.5 & 1 & \( \left(0, \frac{2\pi}{3}, \frac{4\pi}{3}\right) \) \\
    \hline
    noise = True & U(0.1, 1.5) & U(0.1, 1.0) & \( \left(0+\epsilon_1, \frac{2\pi}{3}+\epsilon_2, \frac{4\pi}{3}+\epsilon_3\right) \) \\
    \hline
  \end{tabular}\\[2pt]
  where \( \epsilon_i \sim \mathrm{U}\left(-\frac{\pi}{3.1},\frac{\pi}{3.1}\right) \) and \( z \leftarrow z - \Delta z \).

  \vspace{6pt}
  
  \item Assign the 3 node positions
  \[
  pos_{L,i} = (r\cdot\cos\theta_i,\, r\cdot\sin\theta_i,\, z),\quad i=1,2,3
  \]
  where L indicates the layer and \( i \) indicates the node index within a layer.
\end{enumerate}

\underline{Node Species:}\\[4pt]
if layer < D:\\
    \hspace*{5.3mm} \( x_{L,i} = \mathrm{S}_{L},\; i = 1,2,3 \) \\
else:\\
    \hspace*{5.3mm} \( \texttt{sort\_type} = \texttt{random}(\{\text{ascending, descending}\}) \) \\
    \hspace*{5.3mm} \( \texttt{triplet} = [\mathrm{S}_{\mathrm{D}+1}, \mathrm{S}_{\mathrm{D}+2}, \mathrm{S}_{\mathrm{D}+3}] \) \\
    \hspace*{5.3mm} \( \texttt{order} = \texttt{argsort}(\texttt{triplet}, \texttt{sort\_type}) \) \\
    \hspace*{5.3mm} \( x_{L,i} = \texttt{triplet}_{\texttt{order}_i},\quad i=1,2,3 \)

\vspace{6pt}

\underline{Edges:}\\[4pt]
if layer = 1:\\
    \hspace*{5.3mm} \# Connect chiral center (0) to first layer atoms (1, 2, 3) \\
    \hspace*{5.3mm} \( \texttt{edges.append}([[0, i], [i, 0]]),\quad i=1,2,3 \) \\
else:\\
    \hspace*{5.3mm} \# Connect nodes from the current layer to the layer above \\
    \hspace*{5.3mm} \( \texttt{idx\_prev}, \texttt{idx\_curr} = 3 \times (\text{layer} - 1) + 1,\; 3 \times \text{layer} + 1 \) \\
    \hspace*{5.3mm} \( \texttt{edges.append}([[ \texttt{idx\_prev} + i, \texttt{idx\_curr} + i], [\texttt{idx\_curr} + i, \texttt{idx\_prev} + i]]),\quad i=1,2,3 \) \\

The process will yield a total of 1 + 3D nodes (1 for the chiral center and 3 for each of the D layers).

\subsubsection{Crossed Configurations}
\label{subsubapp:crossed-configurations}
For the \chigeom{} samples of crossed type, each layer 1-D contains a triplet of nodes. The procedure proceeds similarly to the simple samples, but randomizes the edge connections between layers:

\textbf{\underline{Setup:}}\\[4pt]
Identical to simple samples (Appendix~\ref{subsubapp:simple-configurations}).

\textbf{\underline{Chiral Center:}}\\[4pt]
Identical to simple samples (Appendix~\ref{subsubapp:simple-configurations}).

\underline{\textbf{For Layer in range (1, D):}}\\[4pt]
\underline{Positions:}\\[4pt]
Identical to simple samples (Appendix~\ref{subsubapp:simple-configurations}).

\underline{Node Species:}\\[4pt]
Identical to simple samples (Appendix~\ref{subsubapp:simple-configurations}).

\underline{Edges:}\\[4pt]
if layer = 1:\\
    \hspace*{5.3mm} \# Connect chiral center (0) to first layer atoms (1, 2, 3) \\
    \hspace*{5.3mm} \( \texttt{edges.append}([[0, i], [i, 0]]),\quad i=1,2,3 \) \\
else:\\
    \hspace*{5.3mm} \# Connect nodes from the current layer to the layer above \\
    \hspace*{5.3mm} \( \texttt{permutation} = \texttt{permute}([0, 1, 2]) \) \\
    \hspace*{5.3mm} \( \texttt{idx\_prev}, \texttt{idx\_curr} = 3 \times (\text{layer} - 1) + 1,\; 3 \times \text{layer} + 1 \) \\
    \hspace*{5.3mm} \( \texttt{edges.append}([[ \texttt{idx\_prev} + \texttt{permutation}[i], \texttt{idx\_curr} + \texttt{permutation}[i]], \) \newline
    \hspace*{5.3mm} \( [\texttt{idx\_curr} + \texttt{permutation}[i], \texttt{idx\_prev} + \texttt{permutation}[i]]]),\quad i=1,2,3 \) \\

The process will yield a total of 1 + 3D nodes (1 for the chiral center and 3 for each of the D layers).

\subsubsection{Classic Configurations}
\label{subsubapp:classic-configurations}
For the \chigeom{} samples of classic type, each layer 1-D contains a quadruplet of nodes. The procedure goes as follows:

\textbf{\underline{Setup:}}\\[4pt]
We begin by randomly selecting node species for all nodes in the configuration. Specifically, we sample 5 + (D - 1) node species without replacement from the interval [1, species range]. This accounts for one species for the chiral center, four distinct species for the final layer, and D - 1 species for the D - 1 intermediate layers, where each intermediate layer shares the same species across all four of its nodes. Let S be this set of sampled node species and be zero-indexed (i.e. its indices go from zero to D + 4).
\[
\mathrm{S} \sim \text{UniformSubset}((1, \text{species range}), 5+(\mathrm{D}-1))
\]

\textbf{\underline{Chiral Center:}}\\[4pt]
Identical to simple samples (Appendix~\ref{subsubapp:simple-configurations}).

\underline{\textbf{For Layer in range (1, D):}}\\[4pt]
\underline{Positions:}\\[4pt]
If Layer = 1, initialize \( z_{top} = z_{bot} = 1 \).

\begin{enumerate}
  \item Choose parameters

  \vspace{4pt}
  \renewcommand{\arraystretch}{1.5}
  \begin{tabular}{|c|c|c|c|}
    \hline
    \textbf{Noise Flag} & \( \boldsymbol{\Delta z} \) & \( \boldsymbol{r_{bot}}, \boldsymbol{r_{top}} \) & \( \boldsymbol{\theta_{bot}}, \boldsymbol{\theta_{top}} \)\\
    \hline
    noise = False & 0.5 & 1,0 & \( \left(0, \frac{2\pi}{3}, \frac{4\pi}{3}\right), \left(0\right) \) \\
    \hline
    noise = True & U(0.1, 1.5) & U(0.1, 1.0), U(0.0, 1.0) \( \left(0+\epsilon_1, \frac{2\pi}{3}+\epsilon_2, \frac{4\pi}{3}+\epsilon_3\right), \left(\epsilon_4\right) \) \\
    \hline
  \end{tabular}\\[2pt]
  where \( \epsilon_i \sim \mathrm{U}\left(-\frac{\pi}{3.1}, \; \frac{\pi}{3.1}\right), \; i = 1, 2, 3, \; \epsilon_i \sim \mathrm{U}\left(-\pi, \; \pi\right), z_{top} \leftarrow z_{top} + \Delta z, \; \text{ and } z_{bot} \leftarrow z_{bot} - \Delta z \).

  \vspace{6pt}
  
  \item Assign the 4 node positions
    \[
    \begin{aligned}
      pos_{L,i} &= \bigl(r_{\rm bot}\cos\theta_{{\rm bot},i},\;r_{\rm bot}\sin\theta_{{\rm bot},i},\;z_{bot} \bigr), 
        \quad i = 1,2,3,\\[4pt]
      pos_{L,4} &= \bigl(r_{\rm top}\cos\theta_{\rm top},\;r_{\rm top}\sin\theta_{\rm top},\;z_{top} \bigr).
    \end{aligned}
    \]
    where L indicates the layer and \( i \) indicates the node index within a layer.
\end{enumerate}

\underline{Node Species:}\\[4pt]
if layer < D:\\
    \hspace*{5.3mm} \( x_{L,i} = \mathrm{S}_{L},\; i = 1,2,3,4 \) \\
else:\\
    \hspace*{5.3mm} \( \texttt{sort\_type} = \texttt{random}(\{\text{ascending, descending}\}) \) \\
    \hspace*{5.3mm} \( \texttt{triplet} = [\mathrm{S}_{\mathrm{D}+1}, \mathrm{S}_{\mathrm{D}+2}, \mathrm{S}_{\mathrm{D}+3}, \mathrm{S}_{\mathrm{D}+4}] \) \\
    \hspace*{5.3mm} \( \texttt{fourth} = \texttt{min(quadruplet)} \) \\
    \hspace*{5.3mm} \( \texttt{triplet} = \texttt{quadruplet} \setminus \texttt{fourth}  \text{ (the quadruplet without the minimum node species}) \) \\
    \hspace*{5.3mm} \( \texttt{order} = \texttt{argsort}(\texttt{triplet}, \texttt{sort\_type}) \) \\
    \hspace*{5.3mm} \( x_{L,i} = \texttt{triplet}_{\texttt{order}_i},\quad i=1,2,3 \) \\
    \hspace*{5.3mm} \( x_{L,4} = \texttt{fourth} \) \\

\vspace{6pt}

\underline{Edges:}\\[4pt]
if layer = 1:\\
    \hspace*{5.3mm} \# Connect chiral center (0) to first layer atoms (1, 2, 3, 4) \\
    \hspace*{5.3mm} \( \texttt{edges.append}([[0, i], [i, 0]]),\quad i=1,2,3,4 \) \\
else:\\
    \hspace*{5.3mm} \# Connect nodes from the current layer to the layer above \\
    \hspace*{5.3mm} \( \texttt{idx\_prev}, \texttt{idx\_curr} = 4 \times (\text{layer} - 1) + 1,\; 4 \times \text{layer} + 1 \) \\
    \hspace*{5.3mm} \( \texttt{edges.append}([[ \texttt{idx\_prev} + i, \texttt{idx\_curr} + i], [\texttt{idx\_curr} + i, \texttt{idx\_prev} + i]]),\quad i=1,2,3,4 \) \\

The process will yield a total of 1 + 4D nodes (1 for the chiral center and 4 for each of the D layers).

\subsubsection{Final Processing}
\label{subsubapp:final-processing}
After all D layers have been generated, we do the following final processing:
\begin{itemize}
    \item Center: Translate the entire structure so that it is centered at the origin.
    \item Rotate: Apply a uniformly sampled rotation\footnote{We sample our rotation from the set of all 3-dimensional rotations with determinant +1 using SciPy.} that has determinant 1 to the entire structure.
\end{itemize}
These two operations are performed to remove any systematic orientation before the final positions, regardless of the choice of other parameters.

\subsection{Node Labeling}
\label{subapp:node-labeling}
The \ac{cip} rules Section~\ref{subsec:background-chirality} provide a procedure to label chiral centers with exactly four neighbors (as is the case for classic samples), but not for chiral centers with exactly three neighbors (as is the case for simple and crossed samples). To accommodate the simple and crossed sample types while adhering to consistent underlying principles, we mathematically formalize and generalize the \ac{cip} rules. We first define the scalar triple product and show its equivalence to the \ac{cip} rules, then apply the scalar triple product to label other sample types. For \( a_1,a_2,a_3 \in \mathbb{R}^{3} \), the \ac{stp} is defined as
\[
  a_1 \cdot (a_2 \times a_3).
\]
The \ac{stp} is a pseudoscalar, meaning its sign flips under reflections, but does not change under rigid rotations/translations. Since there is a one-to-one correspondence between reflection orientation and chirality, pseudoscalars are natural indicators of chirality.

\subsubsection{Equivalence Between STP and CIP Rules}
\label{subsubapp:stp-cip-equivalence}
In this subsection, we show the equivalence between the clockwise/counterclockwise \ac{cip} labeling described in Section~\ref{subsec:background-chirality} and a certain \ac{stp}. We derive the equivalence below:

\paragraph{Setup} Let \( p_c \) be the position of the chiral center and \( p_1, p_2, p_3, p_4 \) denote the positions of its four substituents, ordered by \ac{cip} priority\footnote{\ac{cip} priority is determined by first ranking from highest to lowest atomic number. If any share the same atomic number, the process continues by comparing additional atomic attributes and, if necessary, recursively comparing attributes of the substituents' neighbors until a difference is found (see Section~\ref{subsec:background-chirality}).} (\( p_1 \) is highest priority, \( p_4 \) is lowest). We prove that there is a 1-to-1 correspondence between the \ac{cip} assignment and the sign of \( (p_4-p_c) \cdot [(p_2-p_1) \times (p_3-p_2)] \).

\paragraph{Orient} Without loss of generality, we translate and rotate the structure so that \( p_c = 0 \) and \( p_4 \) lies along the positive \( z \)-axis, meaning \( p_4 = (0, 0, z^{+}) \) where \( z^{+} \in \mathbb{R}^{+} \). By \ac{cip} convention, the structure is viewed such that the lowest priority substituent is directly behind the chiral center, at a far enough viewpoint to see the entire structure. Since the lowest priority substituent and chiral center are located at \( (0, 0, z^{+}) \) and (0, 0, 0), respectively, the viewpoint must be located \( (0, 0, z^{-}) \), where \( z^{-} \in \mathbb{R}^{-} \) and is sufficiently negative to view the other three neighbors.

\paragraph{Substitute} Plugging \( p_4 \) and \( p_c \) into our \ac{stp} formula, we have:
\[
  (0,0,z^{+}) \cdot [(p_2-p_1) \times (p_3-p_2)].
\]
It is clear that the sign of the \ac{stp} depends solely on the sign of the \( z \)-component of the cross product.

\paragraph{Equivalence} Viewed from the \( (0,0,z^{-}) \) perspective defined earlier, the clockwise / counterclockwise orientation traced by \( p_4 \to p_3 \to p_2 \) also has a one-to-one correspondence with the sign of the \( z \)-component of the cross product \( [(p_2-p_1) \times (p_3-p_2)] \). A positive \( z \)-component indicates a clockwise orientation, while a negative \( z \)-component indicates counterclockwise. Therefore, we have a 1-to-1 relation between the sign of the previously defined \ac{stp} and labeling by \ac{cip} rules, namely:
Positive STP \(\iff\) Clockwise,\\
Negative STP \(\iff\) Counterclockwise.

\subsubsection{Labeling by STP}
\label{subsubapp:stp-labeling}
Having established that the sign of the \ac{stp} provides a formal labeling rule, we now apply it to the other sample types in \chigeom{}. First, we must index the nodal positions to be used in the \ac{stp}. To do so, note that each sample consists of the chiral center and D layers of nodes (see Appendix~\ref{subapp:sample-generation}). Then, index the positions so that \( p_c \) is the position of the chiral center, and \( p_{L,i} \) is the position of the i-th priority node (i = 1 is highest priority) of layer L. We then define the respective \ac{stp} for each sample type:

\[
\begin{aligned}
    \text{Classic:}\quad
    \mathrm{STP} &= (p_{1,4} - p_{c}) \cdot \bigl[(p_{1,2} - p_{1,1}) \times (p_{1,3} - p_{1,2})\bigr],\\[6pt]
    \text{Simple:}\quad
    \mathrm{STP} &= (p_{c} - p_{1,1}) \cdot \bigl[(p_{c} - p_{1,2}) \times (p_{c} - p_{1,3})\bigr],\\[6pt]
    \text{Crossed:}\quad
    \mathrm{STP} &= (p_{c} - p_{\mathrm{D},1}) \cdot \bigl[(p_{c} - p_{\mathrm{D},2}) \times (p_{c} - p_{\mathrm{D},3})\bigr].
\end{aligned}
\]
where D is the chirality distance specified in sample creation (see Section~\ref{sec:chi-geometry}, Appendix~\ref{subapp:sample-generation}).

\subsection{What the Different \chigeom{} Sample Types are Testing}
\label{subapp:sample-types-test-for}
In \chigeom{}, the Classic/Simple samples and the Crossed samples test for different capabilities of the \ac{gnn}, which is now clear given the detailed explanation of the sample creation (Appendix~\ref{subapp:sample-generation}) and node labeling (Appendix~\ref{subapp:node-labeling}). First, note that in all samples, the chirality is determined by the relative (1) geometry and (2) priority of the nodes whose positions are input to the \ac{stp} (call these the \ac{stp} nodes). In classic and simple samples, the \ac{stp} nodes are directly connected to the chiral center, so the relevant geometry is strictly local. However, because all nodes in the intermediate layers share the same node species (by construction), the priority of the \ac{stp} nodes depends on the node species of the nodes in the final layer. Hence, the priority is determined by a variable-range interaction whose hop distance equals the number of layers (see Section~\ref{subsec:background-chirality} for more detail on how the priority is determined). On the other hand, the crossed samples draw all three \ac{stp} nodes from the final layer, which clearly makes both the relative geometry and priority variable-range from the chiral center. Since the number of layers is selected by the user in \chigeom{}, the hop distance of the variable-range interactions is tunable. In summary, the classic and simple samples test a {\ac{gnn}}'s ability to capture short-range geometry and variable-range node attributes, while the crossed samples test a {\ac{gnn}}'s ability to capture variable-range geometry and node attributes.

\subsection{Guaranteeing Chiral Centers in the Samples}
\label{subsec:guaranteeing-chiral-centers}
Each graph constructed by Chi Geometry is engineered to contain exactly one chiral center, labeled R or S. In this section, we show that (i) the initial construction of a sample yields exactly one chiral center and (ii) all subsequent transformations of that sample preserve the labeling of the chiral center.

\subsubsection{Setup}
To be a chiral center, a node must have a certain number of neighbors (3 for simple/crossed samples and 4 for classic), the \ac{cip} ordering of those neighbors must be well-defined, and the requisite \ac{stp} used to assign the label of R or S must be nonzero.

First, note that by construction, each \chigeom{} sample will contain exactly one node with the correct number of neighbors to be a chiral center. Second, see that in \chigeom{} samples, the chiral center connects to a first layer, then each first-layer node connects 1-to-1 to a second layer node, and so on until the final layer. Because every intermediate layer is homogeneous in their node species, the decisive tie-break in \ac{cip} ordering depends on the node species of the final layer (see Section~\ref{subsec:background-chirality} for more detail on \ac{cip} ordering). Since this final layer contains all distinct node species, the ordering is well-defined. Finally, we must show that the respective \ac{stp} of each sample type is nonzero. To show this, we first prove a general case given some assumptions and then show that each of the sample types created by \chigeom{} satisfies those assumptions.

In the general case, assume we have vectors \( v_1,v_2,v_3 \in \mathbb{R}^{3} \), where:
\begin{itemize}
    \item \( v_1 \) contains a nonzero \( z \)-component
    \item \( v_2, v_3 \) lie in the \( xy \)-plane
    \item \(v_2, v_3 \) and are not collinear
\end{itemize}

We will demonstrate that \( \mathrm{STP}(v_1,v_2,v_3) \) is nonzero. First, note that \( (v_2 \times v_3) = (0,0,z_{cross}) \) where \( z_{cross}  \neq 0 \) because \( v_2, v_3 \) lie in the \( xy \)-plane. Then, it clearly follows that \( v_1 \cdot (0,0,z_{cross}) \neq 0 \) because \( v_1 \) has a nonzero \( z \)-component. Thus, \( \mathrm{STP}(v_1, v_2, v_3 ) = v_1 \cdot (v_2 \times v_3 ) \neq 0 \). Now, we show that each sample type fulfills the assumptions above. In order for us to do so, first note these two facts:
\begin{enumerate}
    \item \( \mathrm{STP}(v_1, v_2, v_3) = \det(v_1, v_2, v_3) \)
    \item \( \det(v_1, v_2, v_3) = \det(v_1, v_2+v_1, v_3+v_1) \)
\end{enumerate}

Then, observe the following \acp{stp} for each sample type:
\begin{enumerate}
    \item \textbf{Classic:} \( \mathrm{STP}(p_{1,4}-p_c, p_{1,2}-p_{1,1}, p_{1,3}-p_{1,2}) \).
    \item \textbf{Simple:} \( \mathrm{STP}(p_c-p_{1,1}, p_c-p_{1,2}, p_c-p_{1,3}) = \det(p_c-p_{1,1}, p_c-p_{1,2}, p_c-p_{1,3}) = \det(p_c-p_{1,1}, p_{1,1}-p_{1,2}, p_{1,1}-p_{1,3}) = \mathrm{STP}(p_c-p_{1,1}, p_{1,1}-p_{1,2}, p_{1,1}-p_{1,3}) \).
    \item \textbf{Crossed:} \( \mathrm{STP}(p_c-p_{\mathrm{D},1}, p_c-p_{\mathrm{D},2)}, p_c-p_{\mathrm{D},3}) = \det(p_c-p_{\mathrm{D},1}, p_c-p_{\mathrm{D},2}, p_c-p_{\mathrm{D},3}) = \det(p_c-p_{\mathrm{D},1}, p_{\mathrm{D},1}-p_{\mathrm{D},2}, p_{\mathrm{D},1}-p_{\mathrm{D},3}) = \mathrm{STP}(p_c-p_{\mathrm{D},1}, p_{\mathrm{D},1}-p_{\mathrm{D},2}, p_{\mathrm{D},1}-p_{\mathrm{D},3}) \).
\end{enumerate}

For any of the sample types above, we know that \( v_1 \) has a \( z \)-component and \( v_2, v_3 \) lie in the \( xy \)-plane by construction (see Appendix~\ref{subapp:sample-generation}). Define the positions that are used in \( v_2, v_3 \text{as} P_1, P_2, P_3 \) (there are always 3 distinct positions in any of the sample types). Note that for every sample in \chigeom{} we can label the points such that \( v_2 = P_1-P_2, v_3 = P_1-P_3 \) up to signs, so proving these two difference vectors are non-collinear therefore establishes the result for every sample type. We know by construction that \( P_1, P_2, P_3 \) all lie on a circle of radius \( r > 0 \) (see Appendix~\ref{subapp:sample-generation}). For \( P_1, P_2, P_3 \) to be collinear while lying on the same circle with radius \( r > 0 \), at least two of the points must be equivalent, which implies their angles must be equivalent. But, their angles cannot be equivalent because \( \uptheta = \bigl(0 + \epsilon_{1},\;\tfrac{2\pi}{3} + \epsilon_{2},\;\tfrac{4\pi}{3} + \epsilon_{3}\bigr) \; \text{where} \; \epsilon_{i} \in \bigl(-\tfrac{\pi}{3.1},\,\tfrac{\pi}{3.1}\bigr) \; \text{for} \; i=1,2,3 \). Therefore, \( P_1, P_2, P_3 \) are not collinear. Then, assume for the sake of contradiction that \( P_1, P_2, P_3 \) are not collinear, but \( P_1-P_2 \) and \( P_1-P_3 \) are. Since \( P_1-P_2 \) and \( P_1-P_3 \) are collinear, it follows that for \( c \in \mathbb{R}, \; c(P_1-P_2) = P_1-P_3 \to P_3=[(1-c)P_1] + cP_2 \). But, this contradicts the fact that \( P_1, P_2, P_3 \) are not collinear. Therefore, by contradiction, \( P_1-P_2 \) and \( P_1-P_3 \) must not be collinear.

\subsubsection{Transformations}
In this subsection, we show that \( \det(v_1, v_2, v_3) \) (which is equal to \( \mathrm{STP}(v_1, v_2, v_3) \)) remains unchanged by the transformations after the initial sample construction. Since the \ac{stp} is used to assign chirality labels, this implies that the transformations do not corrupt the label assignment. First, note that:
\begin{itemize}
    \item For every sample type in \chigeom{}, all \( v_i \; (i=1,2,3) \) in that sample type's respective \ac{stp} are relative vectors.
    \item For any two square matrices \( \mathrm{A},\mathrm{B} \in \mathbb{R}^{n \times n}, \; \det(\mathrm{AB}) = \det(\mathrm{A})\det(\mathrm{B}) \)
\end{itemize}

Then, for the two transformations (centering + rotation):
\begin{enumerate}
    \item \textbf{Centering:} translate the structure by \( t \in \mathbb{R}^{3} \) so that the structure is centered at the origin.
    \begin{itemize}
        \item The determinant is unchanged by translation of the structure because relative vectors are unchanged by translations.
    \end{itemize}
    \item \textbf{Rotation:} apply a random rotation \( \mathrm{A} \in \mathbb{R}^{3 \times 3} \; \text{where} \; \det(\mathrm{A}) = 1 \).
    \begin{itemize}
        \item Rotating the positions will correspondingly rotate the relative vectors. Denote \( \det(v_1, v_2, v_3) \; \text{as} \; \det(\mathrm{V}) \; \text{and} \; \det(\mathrm{A}v_1, \mathrm{A}v_2, \mathrm{A}v_3) \; \text{as} \; \det(\mathrm{AV}) \). Then, the determinant of the rotated structure \( \det(\mathrm{AV}) = \det(\mathrm{A}) \cdot \det(\mathrm{V}) = \det(\mathrm{V}) \). Therefore, the determinant is unchanged by rotation.
    \end{itemize}
\end{enumerate}

Therefore, the \ac{stp} (and equivalently the chirality labeling) is unchanged by the centering and random rotation applied after initial sample construction. 

\section{Equivariance}
\label{app:equivariance}

\paragraph{Definition} Equivariance is a mathematical concept in group and representation theory, with strong applications for ML models in molecular, materials, and biological prediction tasks. We use the following definition of equivariance by \citet{equivariance_kondor}:

Let G be a group and \(X_{1},X_{2}\) be two sets with corresponding G-actions
\[
T_{g}:X_{1}\to X_{1}, \quad T'_{g}:X_{2}\to X_{2}.
\]
Let \( V_{1} \) and \( V_{2} \) be vector spaces, and \( \mathbb{T}, \; \mathbb{T'} \) be the induced actions of G on \( L_{V_{1}}(X_{1}) \) and \(L_{V_{2}}(X_{2}) \).

We say that a map \( \phi: L_{V_{1}}(X_{1}) \to L_{V_{2}}(X_{2}) \) is equivariant with the action of G (or G-equivariant for short) if \( \phi(\mathbb{T}_g f) = \mathbb{T'_g}\phi(f) \quad \forall f \in  L_{V_{1}}(X_{1}) \) for any group element \( g \in \mathrm{G} \). 

Succinctly, equivariant means that transformations and mappings commute. Invariance is the special case where \( \mathbb{T'_g} = \mathrm{id}_{L_{V_{2}}(X_{2})} \) (the identity). In other words, applying the G-action does not change the output of \( \phi \).

Additionally, we define an equivalence class under the G-action as the set of all elements that can be mapped to one another by transformations in G. Any single element from this set may serve as its representative.

\paragraph{Examples} Some real-world examples of equivariance in chemistry and materials science include:
\begin{enumerate}
    \item \textbf{Free Energy:} Free energy remains invariant under any rotation, translation, or reflection.
    \item \textbf{Interatomic Forces:} Interatomic forces rotate correspondingly with the structure, making them equivariant to rotations.
    \item \textbf{Chiral Center R/S Label:} A chiral center's tag (R or S) is invariant to translations and rotations, and equivariant to reflections.
\end{enumerate}

Impact When configured correctly, equivariant models inherently respect the symmetries of their prediction task/s. By treating all elements in an equivalence class as transformed versions of a single representative, these models reduce the dimensionality of the functional space in which they learn. This translates into an easier learning task, resulting in improved performance and learning efficiency \citep{alphafold2_jumper, gnome_merchant, nequip2022, mace2022, gcpnet2022}.

\section{Over-squashing Analysis}
\label{app:oversquashing}

The following method is used to perform an analysis of over-squashing in the experiment from Section~\ref{subsec:benchmarking-existing-architectures}. In this experiment, we trained the model on nine distinct datasets parametrized by chirality distance \( \mathrm{D} \in \{1,2,\dots,9\} \). The chirality distance parameter controls how many connection `hops' separate the chiral center from the substituent nodes which determine its chiral tag. Once each model is trained, we analyze over-squashing on its corresponding dataset by measuring the gradients of node features with respect to the chiral center classification loss.

The over-squashing analysis proceeds in three main steps:\\[4pt]
\textbf{1. BFS for Hop Distances}\\[2pt]
Let c be the chiral center in the graph G = (V, E). For each node \( v \in \mathrm{V} \), define its hop distance \( f(v) \) to c as the length of the shortest path from c to \( v \). For example, \( f(\mathrm{c}) = 0 \), \( f(v) = 1 \) if v is directly connected to c, and so on.

\textbf{2. Gradient Calculation}
Let CE be the cross-entropy classification loss computed only on the chiral center c. Each node \( v \) has a feature vector \( x_v \). We backpropagated on CE to find \(\mathrm{CE}\) to find \(\nabla_{x_v}\mathrm{CE}\), then measure node \( v \)'s importance via its respective gradient's \( L_2 \)-norm:
\[
  g(v) = \bigl\|\nabla_{x_v}\mathrm{CE}\bigr\|_2.
\]
We collected this value across all nodes for all graphs in the dataset.

\textbf{3. Aggregation and Normalization}
For each distance \( d \in \{1,2,\dots,\mathrm{D}\} \)
, we averaged \( g(v) \) over all nodes \( v \) with \( f(v) = d \). Then, we normalized these averages so they sum to 1:
\[
  g(d)
  = \frac{1}{\bigl|\{\,v : f(v)=d\}\bigr|}\,
    \sum_{f(v)=d} g(v),
  \qquad
  \hat g(d)
  = \frac{g(d)}{\sum_{d=1}^{\mathrm{D}} g(d)}.
\]
The value of \( \hat{g}(d) \) indicates the relative importance of nodes at hop distance d from the chiral center. For any \chigeom{} dataset generated with chirality distance D, the nodes exactly D hops away from the chiral center determine the center's R/S label. Therefore, a well-trained model should assign these deciding nodes a large share of the gradient. If the relative gradient of the deciding nodes (represented by \( \hat{g}(\mathrm{D}) \) plummets as D grows, it indicates that over-squashing is taking place.

\section{Computational Cost and Hardware}
\label{app:compute}

Wall-clock durations were computed as the difference between the configuration save timestamp (immediately before training) and the final-results timestamp (immediately after training/evaluation). All runs were executed on a MacBook Pro (14 inch) with an Apple M3 Pro CPU and 36 GB of RAM. The total runtime across all experiments is \textbf{313.16 hours}.

\begin{table}[h!]
  \centering
  \caption{Training durations (decimal hours) for \( \mathrm{D}=1 \dots 9 \).}
  \label{tab:compute-timestamps}
  \begin{tabularx}{\linewidth}{l *{10}{>{\centering\arraybackslash}X}}
    \toprule
    \textbf{Model} 
      & \textbf{D=1} & \textbf{D=2} & \textbf{D=3} & \textbf{D=4}
      & \textbf{D=5} & \textbf{D=6} & \textbf{D=7}
      & \textbf{D=8} & \textbf{D=9} & \textbf{Total} \\
    \midrule
    E3NN Virtual Node
      & 1.40 & 1.42 & 4.78 & 4.62 & 1.75 & 4.80 & 2.28 & 2.55 & 3.32 & 26.92 \\

    \addlinespace
    Global E3NN (Feat.\ Eng.)
      & 2.18 & 3.65 & 4.92 & 5.60 & 19.00 & 20.65 & 27.08 & 50.98 & 59.00 & 193.23 \\

    \addlinespace
    Local E3NN
      & 3.23 & 5.13 & 5.43 & 8.30 & 9.60 & 3.90 & 4.15 & 3.58 & 4.10 & 47.42 \\
    \bottomrule
  \end{tabularx}
\end{table}

\begin{table}[h!]
  \centering
  \caption{Training durations (decimal hours) for repetitions \( \mathrm{R}=1 \dots 10 \) at \( \mathrm{D}=1 \).}
  \label{tab:compute-reps}
  \begin{tabularx}{\linewidth}{%
      l*{10}{>{\centering\arraybackslash}X}
      >{\centering\arraybackslash}X
      >{\centering\arraybackslash}X
    }
    \toprule
    \textbf{Model}
      & \textbf{R=1} & \textbf{R=2} & \textbf{R=3} & \textbf{R=4}
      & \textbf{R=5} & \textbf{R=6} & \textbf{R=7} & \textbf{R=8}
      & \textbf{R=9} & \textbf{R=10}
      & \textbf{Total} \\
    \midrule
    Global E3NN
      & 1.02 & 0.98 & 1.12 & 1.22 & 1.90 & 2.20 & 2.27 & 1.57 & 2.33 & 2.23
      & 16.84 \\

    SE(3)-Transformer
      & 2.97 & 3.07 & 2.98 & 2.65 & 2.57 & 2.55 & 2.57 & 2.55 & 2.57 & 2.55
      & 27.03 \\

    Vanilla MPNN
      & 0.13 & 0.13 & 0.12 & 0.13 & 0.13 & 0.40 & 0.15 & 0.15 & 0.23 & 0.15
      &  1.72 \\
    \bottomrule
  \end{tabularx}
\end{table}

\end{document}